# PointNet++ Grasping: Learning An End-to-end Spatial Grasp Generation Algorithm from Sparse Point Clouds

Peiyuan Ni[1], Wenguang Zhang[1], Xiaoxiao Zhu[1], Qixin Cao[1]

*Abstract*— Grasping for novel objects is important for robot manipulation in unstructured environments. Most of current works require a grasp sampling process to obtain grasp candidates, combined with local feature extractor using deep learning. This pipeline is time-costly, especially when grasp points are sparse such as at the edge of a bowl.

In this paper, we propose an end-to-end approach to directly predict the poses, categories and scores (qualities) of all the grasps. It takes the whole sparse point clouds as the input and requires no sampling or search process. Moreover, to generate training data of multi-object scene, we propose a fast multi-object grasp detection algorithm based on Ferrari Canny metrics. A single-object dataset (79 objects from YCB object set, 23.7k grasps) and a multi-object dataset (20k point clouds with annotations and masks) are generated. A PointNet++ based network combined with multi-mask loss is introduced to deal with different training points. The whole weight size of our network is only about 11.6M, which takes about 102ms for a whole prediction process using a GeForce 840M GPU. Our experiment shows our work get 71.43% success rate and 91.60% completion rate, which performs better than current state-of-art works.

## I. INTRODUCTION

Object manipulation in unstructured environments of real world is still an open problem, especially for unseen objects. However, many high challenging problems still exist: 1) When objects are stacked in a pile, it is difficult and time-costly to search available grasps. 2) The camera data may be sparse and noisy, which makes it challenging to generate 3D spatial grasps. 3) Appropriate quality metrics should be considered to obtain a best grasp among all the grasp candidates.

Grasping preception adopts different algorithms depending on the characteristics of the grasping scene [1]. Traditional model-based algorithms apply 6D pose estimation algorithms [2], [3] to obtain object poses and choose a best grasp from a pre-built grasp database. This database can be predefined manually or generated by other tools such as Graspit! [4]. However, this pipeline can hardly be applied to novel objects. Another approach is to apply learning algorithm to learn grasp representations. In this aspect, most of works use a supervised framework that the grasps are labeled manually [5] or automaticly at first [6-8]. Then these works usually utilize different data forms, such as depth image [6], RGB-D image [5] and point clouds [8], to represent the sensor information for a grasp candidate. And then deep learning algorithms, such as convolutional neural networks (CNNs) and PointNet [9], are applied to extract the features of these grasps. However, this pipeline only focuses on local features around a grasp, which is unable to be combined with global information such as the distribution of objects. Some other works utilize CNNs [10] and fully convolutional networks (FCNs) [11] to train a grasp metric distribution and the whole grasp scene is regarded as the input. Moreover, other researchers even treat this problem as a Markov Decision Process (MDP) problem, and an end-to-end learning process is trained in real world directly [12], [13] or by simulation learning [14], [15] to guide the grasping action. These methods perform well even in dense-cluttered scene with only one RGB camera. However, the whole training process, especially the feature extraction part, is still specialized for 2D or 2.5D (depth map) grasping. For 3D spatial grasping, a huge quantity of data are required during reinforcement learning process in order to get a high precision. Moreover, traditional feature extractor, such as FCNs and CNNs, is not suitable for 3D tasks such as point classification and scene understanding [16].

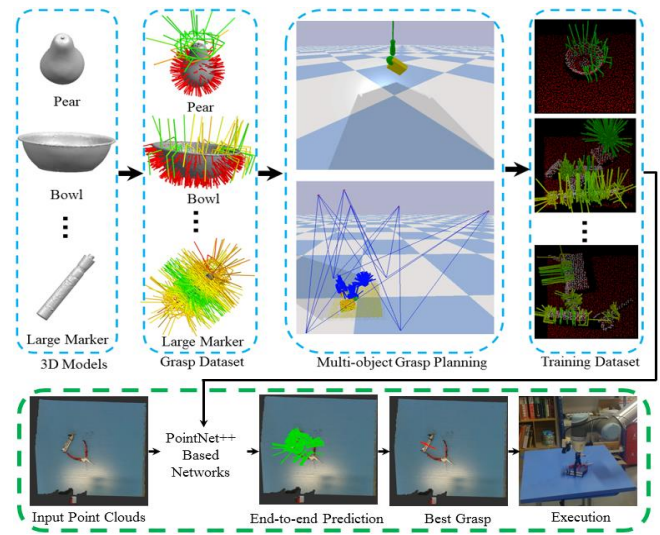

Figure 1. The pipeline of our work. Our synthetic training dataset is based on YCB [26] object set. Given raw sensor data from a RGB-D camera, our PointNet++ based Network can directly predict the poses, categories and scores (qualities) of all the grasps in a fast way.

Therefore, to tackle these problems, we propose an end-to-end network based on PointNet++ [16] to directly predict the grasps and their qualities, shown in Figure 1. Compared to several similar works [6-8], our algorithm abandons traditional learning pipeline and takes the whole point clouds as input to regress the spatial grasp poses. In this way, the prediction process can be combined with global data information. Moreover, it requires no grasp sampling process

[1]State Key Lab of Mechanical Systems and Vibration.
Intelligent Robot Laboratory of SJTU, Shanghai Jiao Tong University, Shanghai, China

{pyni_sjtu, zhwg, ttl, qxcao}@sjtu.edu.cn

[6-8] and every point can predict a result, which saves huge sampling costs especially when grasp points are sparse. The whole weight size of our network is only about 11.6M, which takes about 102ms for a whole prediction process using a GeForce 840M GPU. Another key point is how to generate the training database with grasping metric for our algorithm, especially for large cluttered scenes with multiple objects. Many database generation processes for spatial grasps are based on single objects [7], [8], [10]. Some works [6], [18] have considered cluttered scenes, but these methods are still time-consuming. Exhaustive search direcly in a large scene takes huge costs and has low efficiency. We propose to search grasp candidates with Ferrari Canny metrics [17] for each single object at first. Then these candidates are matched in the same object for a multi-object scene. Therefore, for each large scene, only collision detection is required. However, this brings a problem that how to deal with different grasp points with different labels. We introduce a multi-mask loss to deal with this problem, the detail of which will be explained in training part. To summarize, the contributions of this paper are shown as follows:

1) An end-to-end network based on PointNet++ [16] combined with a multi-mask loss is proposed to directly predict poses, categories and scores (qualities) of spatial grasps. It takes the whole sparse point clouds as the input and requires no sampling or search process.

2) A fast multi-object grasp detection algorithm based on Ferrari Canny metrics [17] is proposed. Based on single-object dataset (79 objects from YCB [26], 23.7k grasps) generated by our single-object grasp planning, a large-scale grasp dataset for multiple objects using domain randomization is built, which contains 20k point clouds with annotations and masks. The dataset is published in: https://github.com/pyni/PointNet2_Grasping_Data_Part.

## II. RELATED WORK

For grasping preception, the most important factor that influences how to choose the algorithm is the prior object knowledge [1]. For unknown objects, on the one hand, deep learning techniques have been widely used in this case. A better network structures, such as CNNs [5-7] and FCNs [11], help to extract better features and avoid exhaustive search. Our previous work also focuses on this aspect [28]. However, a training pipeline (sampling a grasp, labeling the grasp and extracting the features of the grasp) is unavoidable for most of these works. This brings two problems: 1) In test time, a grasp sampling process is required. When grasp points are sparse, it takes huge sampling costs in order to find them out. 2) The feature extractor is based on 2D representation learning, which is not suitable to deal with sparse 3D point clouds and spatial grasps. Meanwhile, a grasping process can also be considered as a markov decision process (MDP) [12]. Many deep reinforcement learning techniques [12], [13], [20] are competent for this, which makes the whole grasping process in an end-to-end way from preception to control. However, for spatial grasps, huge costs are required to train a high precision with deep reinforcement learning. And it is not general for different robots.

With the development of deep learning [9], [16] on 3D data, the feature extractor is promoted from 2D representation learning into 3D geometric representation learning. Many tasks based on 3D data, such as segmentation, recognition and correspondence, can be realized [21]. In [19], researchers utilize a three-dimensional deep convolutional neural network (3D CNN) to deal with the voxeled point clouds. However, many details of original point clouds are lost during voxelization. Referring the work in [7], the authors of [8] propose to use PointNet [9] as the feature extractor to deal with original point clouds. However, this work is still unable to avoid the problem of this kind of training pipeline [5-8]: Grasp sampling process is required in test time. The final performances (grasp precision and time cost) rely heavily on sampling method. Although some sampling methods such as cross-entropy method (CEM) [6], [12], [22] can accelerate search speed, the problem can't be solved essentially, the precision of which depends on iteration time. Moreover, if grasp points are sparse in a large scene, more iteration time is required to find them out. Therefore, we abandon this idea and consider an end-to-end pipeline to regress spatial grasps.

## III. PROBLEM STATEMENT

### A. Assumptions

Our grasping model is based on the following assumptions: 1) quasi-static physics with Coulomb friction, 2) two-fingered parallel-jawed grasping with known geometry parameters, and 3) one depth camera with known intrinsics.

For spatial grasps, some approaches [5], [7], [8], [27], [28] prefer to choose the grasp, approaching direction of which is along surface normal of the objects. This is also the preference of humans. Therefore, in order to simplify the grasping model, the following assumption is added: 4) approaching direction of a grasp is along surface normal of the objects.

### B. Definitions

Some definitions are introduced here.

**Object States.** Let $x=(x_1, x_2...x_m)$ denote a state describing a grasping scene that containing $m$ objects. $x_i=(\mathcal{O}_i, M_i, C_i, T_i^o, \mu_i)$ represents the sub-state for object $i$, where $\mathcal{O}_i$ specifies the surface model of object, $M_i, C_i, T_i^o$ are the mass, centroid and pose of the object $i$, respectively, and $\mu_i \in \mathbb{R}$ is the coeffeicient of friction between the object $i$ and gripper.

**Camera State.** Let $c=(T^c, i)$ denote the camera state with camera pose $T^c$ and intrinsics parameters $i$.

**Point Clouds.** Let $y \in \mathbb{R}^{3\times N}$ denote the point clouds that contains N points by the depth camera.

**Grasps.** Let $g=(g_1, g_2...g_n)$ denote grasps in 3D space. We define each grasp further: $g_i=(p, n, r, d), j=1,2...n$. Here $p=(p_x, p_y, p_z) \in \mathbb{R}^3$ is grasp point, which is to locate the position of the grasp. Usually grasp point $p$ is on the point clouds $y$. $n=(n_x, n_y, n_z) \in \mathbb{R}^3$ and $r=(r_x, r_y, r_z) \in \mathbb{R}^3$ are the unit approaching vector and opening vector, respectively, which are shown in Figure 2. Based on assumption 4), $n$ actually is the normal of grasp point $p$. Here $p, n, r$ are relative to the absolute coordinate system of point clouds $y$. $d \in \mathbb{R}$ is the approaching distance of two fingers relative to grasp point $p$ along approaching vector $n$.

**Grasp Metircs.** Let $Q_{fc}(g_i) \in [0,1]$ be the continuous grasp metircs for grasp $g_i$. Because regression task is more difficult to be trained compared to classification task, we set a binary-valued grasp metircs for grasp $g_i$, which can be denoted as $Q_b(g_i) \in \{0,1\}$. In our training stage, the relationship between $Q_b(g_i)$ and $Q_{fc}(g_i)$ is defined as follows:

$$Q_b(g_i)=\begin{cases} 0 & \text{if } Q_{fc}(g_i)==0 \\ 1 & \text{if } Q_{fc}(g_i)>0 \end{cases} \quad (1)$$

Furtherly, the combined metrics is defined:

$$Q_c(g_i)=1\{Q_b(g_i)==1\}\cdot Q_{fc}(g_i) \quad (2)$$

, where $1\{\cdot\}$ is the indicator function.

Given point clouds $y$ of object states $x$ and camera state $c$, our goal is to learn a function $\mathcal{F}(y)$, which maps $y$ into all the grasps $g$ and their grasp metrics $Q_c(g)$. The grasp with maximum $Q_c$ is executed by the robot.

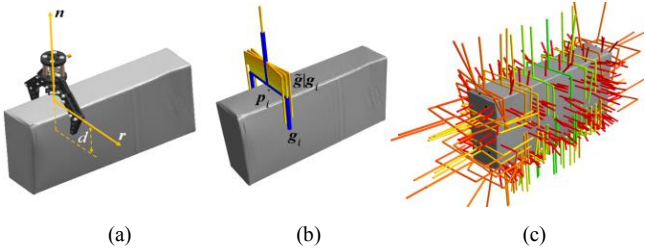

(a)        (b)        (c)

Figure 2. (a) Grasping model. (b) Major grasp $g_i$ (the grasp with thick blue lines) and supplementary grasps $\tilde{g}|g_i$ (the rest grasps). (c) An example of $n_{neg}$ and $g$ with $Q_c(g)$. The colors from red to green represent $Q_c(g)$ from low to high. $n_{neg}$ are represented by red lines.

## IV. DATA GENERATION

### A. Single-object Grasp Planning

Given an object model $\mathcal{O}_i$ with initial values ($M_i$, $C_i$, $T_i^o$, $\mu_i$), the target of single-object grasp planning is to obtain grasps $g$, metrics $Q_c(g)$, supplementary grasps $\tilde{g}$ and negative grasp points and normals ($p_{neg}$, $n_{neg}$). Some definitions will be introduced later. According to Ferrari Canny metrics [17], the forces and torques of the gripper can be represented by points in the wrench space $\mathcal{W}$, $\mathcal{W} \in \mathbb{R}^6$. If a grasp $g_i$ is force-closure, the origin of wrench space $\mathcal{W}$ is contained in the convex hull of the primitive wrenches [24], [17] (, which is denoted as $\mathcal{BG}$). Moreover, the grasp quality metrics $Q_{fc}(g_i)$ is the distance between the origin and the nearest point in the boundary of $\mathcal{BG}$:

$$Q_{fc}(g_i)=\min_{\omega \in \text{Bd}(\mathcal{BG})} \|\omega\| \quad (3)$$

, where $\text{Bd}(\mathcal{BG})$ represents the boundary of $\mathcal{BG}$.

At first, given an object model $\mathcal{O}_i$, N points will be randomly sampled and their normals $n$ are also calculated. With assumption 4), a grasp is generated from each sampling point $p_i$ and normal $n_i$. We rotate it for $N_r$ times along approaching vector $n$. For each rotation, the gripper $\mathcal{O}_g$ will do the collision detection from a far distance to the deepest approaching distance $d_{deepest}$. If $d_{deepest}$ is larger than $D_{min}$, this rotation is considered as valid and added into a set denoted as $g^r$.

If $g^r$ is an empty set, it means no grasp can be generated from $p_i$. Then we regard $p_i$ and $n_i$ as negative grasp points and normals, denoted by $p_{neg}$ and $n_{neg}$. Otherwise, for each grasp $g_j^r \in g^r$, the combined metrics $Q_c(g_j^r)$ will be calculated using equation (2) and (3). If $Q_c(g_j^r)>0$, it means $g_j^r$ is graspable. Finally, the metrics for all the grasps in $g^r$ will be calculated. The grasp with maximum metrics value is considered as a *major grasp* and is added into $g$.

Because in multi-object grasp planning, a major grasp may be unreachable due to occlusion or collision, it is necessary to record the supplementary grasps $\tilde{g}$ for each major grasp. To obtain $\tilde{g}$, we sample and collect grasps every 1 mm from $D_{min}$ to $d$. $Q_c(\tilde{g})$ of supplementary grasps $\tilde{g}$ are also calculated using equation (2) and (3). Then grasps $\tilde{g}$ with $Q_c(\tilde{g})=0$ are filtered, and the remaining supplementary grasps are sorting in descending order according to $Q_c(\tilde{g})$. An example to describe $g_i$ and its supplementary grasps $\tilde{g}|g_i$ in $p_i$ are shown in Figure 2 (b).

$\mu$, N, $N_r$ and $D_{min}$ are set to 0.2, 300, 36 and 20 mm in our experiment. Centroid $C_i$ is calculated by shape distribution of the model. Mass $M_i$ is calculated by multiplying model's volume by density (, which is set to 0.5 g/cm$^3$ in our experiment). Finally, for all the training objects (, which are denoted as $\mathcal{O}_1, \mathcal{O}_2, \mathcal{O}_3...\mathcal{O}_n$), the outputs are denoted as $\{g, Q_c(g), \tilde{g}, Q_c(\tilde{g}), (p_{neg}, n_{neg})| \mathcal{O}_1, \mathcal{O}_2, \mathcal{O}_3...\mathcal{O}_n\}$. Figure 2 (c) shows an example of $n_{neg}$ and $g$ with $Q_c(g)$.

### B. Multi-object grasp planning

The objects may be stacked in a pile in our test scene. To simulate this case and generate training data, a multi-object grasp planning algorithm (Algorithm 1) is introduced. The simulator we used is Pybullet [25].

---

**Algorithm 1** Multi-object grasp planning

Input: Object models $\mathcal{O}_1, \mathcal{O}_2, \mathcal{O}_3...\mathcal{O}_n$, Gripper model $\mathcal{O}_g$, Outputs of single-object grasp planning for all the objects $\{g, Q_c(g), \tilde{g}, Q_c(\tilde{g}), (p_{neg}, n_{neg})| \mathcal{O}_1, \mathcal{O}_2, \mathcal{O}_3...\mathcal{O}_n\}$, Query radius R

Output: Training data $y_t$, Training labels $l_t$, Training masks $m_t$

1: **define** $p_{pos}=[\ ], g_{pos}=[\ ], \tilde{p}_{neg}=[\ ], \tilde{n}_{neg}=[\ ], y_t=[\ ], l_t=[\ ]$
2: $\mathcal{O}_{r_1}, \mathcal{O}_{r_2}...\mathcal{O}_{r_m}$=Randomly_ selection($\mathcal{O}_1, \mathcal{O}_2, \mathcal{O}_3...\mathcal{O}_n$)
3: $\mathcal{O}_{scene}$=Simulator_initialization($\mathcal{O}_{r_1}, \mathcal{O}_{r_2}...\mathcal{O}_{r_m}$)
4: **for all** $\mathcal{O}_{r_i} \in \{\mathcal{O}_{r_1}, \mathcal{O}_{r_2}...\mathcal{O}_{r_m}\}$ **do**
5:      **for all** $g_j \in \{g|\mathcal{O}_{r_i}\}$ **do**
6:          **if** Collision_detection($\mathcal{O}_{scene}, g_j$) == False:
7:              add $p|g_j$ to $p_{pos}$, add $g_j$ to $g_{pos}$
8:          **else**:
9:              $g^*$=Collision_detection_with_ supplementary_grasps ($\tilde{g}|g_j, Q_c(\tilde{g}|g_j)$)
10:            **if** $g^*$ == $\varnothing$ **then**:
11:                add $p|g_j$ to $\tilde{p}_{neg}$, add $n|g_j$ to $\tilde{n}_{neg}$
12:            **else:**
13:                add $p|g_j$ to $p_{pos}$, add $g^*$ to $g_{pos}$
14:      **end for**
15: **end for**
16: $y, l, m$ =Camera_state_randomization()

17: $y_t, l_t, m_t$ =Labels_generation($y, l, m, p_{pos}, g_{pos}, \tilde{p}_{neg}, p_{neg}, \tilde{n}_{neg}, n_{neg}, R$ )

In step 2, given object models $\{\mathcal{O}_1, \mathcal{O}_2, \mathcal{O}_3 \ldots \mathcal{O}_n\}$, $m$ models are randomly chosen: $\mathcal{O}_{r_1}, \mathcal{O}_{r_2} \ldots \mathcal{O}_{r_m}$. Then in simulator, these models are randomly chosen and thrown in an invisible totebox in step 3, which is to limit the range of the whole scene into 200mm×200mm. The whole scene including objects and the ground plane is considered as a whole model: $\mathcal{O}_{scene}$.

For each grasp $g_j$ of each object $\mathcal{O}_{r_i}$ obtained by single-object grasp planning, collision detection is done between $g_j$ and $\mathcal{O}_{scene}$ in step 6. If no collision occurs, $g_j$ and its grasp point $p$ are stored into $g_{pos}$ and $p_{pos}$. Otherwise, all the supplementary grasps $\tilde{g}$ of $g_j$ will be applied to do the collision detection in step 9. The collision-free grasp with maximum quality value $Q_c(\tilde{g}|g_j)$ is selected and denoted as $g^*$. If $g^*$ is an empty set, which means no grasp is generated from this point, the grasp point and normal of $g_j$ are collected into $\tilde{p}_{neg}$ and $\tilde{n}_{neg}$ in step 11. Otherwise, $g^*$ and its grasp point are collected into $g_{pos}$ and $p_{pos}$.

From step 4 to step 15, our algorithm is to make a multi-object grasp planning. The next stage is to generate training data and labels. In step 16, the pose of camera is randomized under a constraint that optical axis of the camera should pass through the center $o_w$ of the invisible totebox, shown in Figure 3. The resolution of camera in simulation is 640×480. The figure in the middle of Figure 1 shows five random poses of the camera.

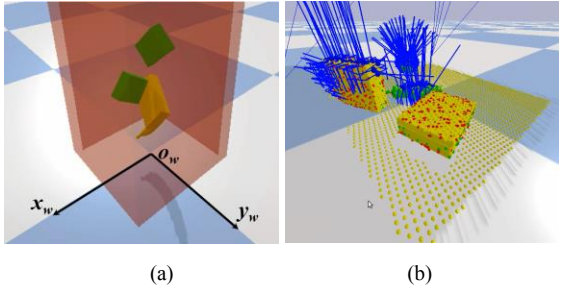

(a)          (b)

Figure 3. (a) The invisible totebox. The coordinate $x_w$-$o_w$-$y_w$ is the absolute coordinate system.(b) Different training points. The yellow points are obtained by camera within 200mm×200mm. The red, green and blue points are obtained by step 4 to step 15, where the red points are $p_{neg}$, the green points are $\tilde{p}_{neg}$ and blue points are $p_{pos}$.

Then the point clouds $y$ within 200mm × 200mm are obtained. Because $y$ is not matched with $p_{pos} \cup \tilde{p}_{neg} \cup p_{neg}$, we should introduce masks to define and distinguish different points. Here the label and mask for each point of $y$ is defined as follows:

Label for each point = $[n_x, n_y, n_z, r_x, r_y, r_z, Q_b(g_i), Q_c(g_i)]$

Mask for each point = $[m^i_{normal}, m^i_{rotation}, m^i_{category}, m^i_{score}]$

To simplify work, we use $\mathbf{L}^i_{normal}, \mathbf{L}^i_{rotation}, L^i_{category}$ and $L^i_{score}$ to represent $(n_x, n_y, n_z)$, $(r_x, r_y, r_z)$, $Q_b(g_i)$ and $Q_c(g_i)$:

Label for each point = $[\mathbf{L}^i_{normal}, \mathbf{L}^i_{rotation}, L^i_{category}, L^i_{score}]$

The masks and labels for the whole point clouds $y$ are denoted as $m$ and $l$. With prior knowledge, we can set some initial values of $m$ and $l$ for $y$. Specifically, the labels $l$ and masks $m$ of point clouds on the ground are set to [$\mathbf{0, 0}$, 0, 0] and [0, 0, 1, 0], which means only $Q_b$ of these points are considered during training stage. Similarly, the labels $l$ and masks $m$ of point clouds on the objects are set to [$\mathbf{0, 0}$, 0, 0] and [0, 0, 0, 0].

In step 17, $y$, $l$ and $m$ are added into $y_t$, $l_t$ and $m_t$ at first. Then for each point $p_i \in p_{pos}$, a kd-tree search is applied to find the nearest point among $y$ with query radius of R. If the nearest point is found, $p_i$ is added into $y_t$. Moreover, its label and mask are set to $[n|g_i, r|g_i, 1, Q_c(g_i)]$ and [1, 1, 1, 1], where $g_i \in g_{pos}$. For each point $p_i \in \tilde{p}_{neg} \cup p_{neg}$, the similar process is done. The difference is that label and mask for $p_i$ are set to $[n_i, \mathbf{0}, 0, 0]$ and [1, 0, 1, 1], where $n_i \in \tilde{n}_{neg} \cup n_{neg}$. An example is given in Figure 3 to show different points obtained from step 16 to step 17. The whole algorithm only makes one simulation, which provides $N_{cam}$ cases, where $N_{cam}$ denotes the number of camera shots for each simulation. Finally, $N_{sim} \cdot N_{cam}$ samples ($y_t, l_t, m_t$) are generated totally, where $N_{sim}$ denotes the number of simulations.

$N_{sim}$, $N_{cam}$ and R are set to 4000, 5 and 1 mm in our experiment. $m$ is selected randomly from the set of integers between 1 and 8 for each simulation. Figure 4 shows some examples of our generated training points.

## V. TRAINING

For $y_t \in \mathbb{R}^{3 \times N'}$, the point clouds are centralized to zero mean. All the training data are scaled with a factor $M$, which makes sure that all the training data are scaled into [0, 1]. $M$ is set to 0.33 in our experiment. To simulate the noise, the position of each points is jittered by a gaussian noise with zero mean and 1 mm standard deviation. Then the training data are resampled into 8192 points, which is denoted as $\hat{y}_t \in \mathbb{R}^{3 \times 8192}$. The network structure is shown as follows:

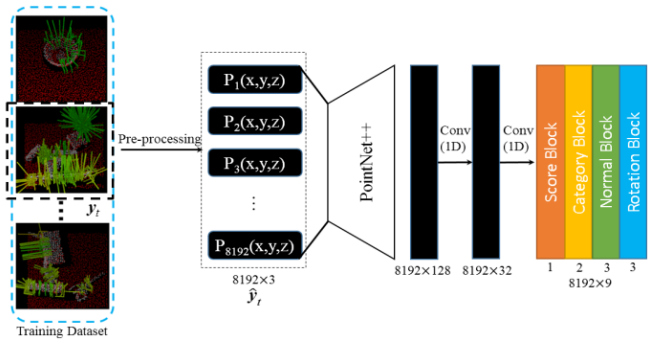

Figure 4. Generated training points and network structure of our algorithm. The red points are points with $m^i_{category}$ and $L^i_{category}$ equal to 1 and 0. The white points are points with $m^i_{category}$ equal to 0. Grasps are plotted at the points with $m^i_{category}$ and $L^i_{category}$ equal to 1 and 1.

A PointNet++ [16] based network followed by two one-dimensional convolution filters is applied to deal with the processed point clouds $\hat{y}_t$. The parameters of PointNet++ parts are shown as follows:

SA(1024,0.1,[32,32,64]) → SA(256,0.3,[64,64,128]) → SA(64,0.5,[128,128,256]) → SA(16,1.0,[256,256,512]) → FP(256,256) → FP(256,256) → FP(256,128) → FP(128,128,128).

Defined by [16], SA($K$, $r$, $[l_1,...,l_d]$) is a set abstraction (SA) level with $K$ local regions of ball radius $r$ using PointNet of $d$ fully connected layers with width $l_i$ ($i$ = 1, ..., $d$). FP($l_1,...,l_d$) is a feature propagation (FP) level with $d$ fully connected layers.

The output is a feature block with size of 8192×9. Then the output feature is divided into four parts: score block (8192×1), category block (8192×2), normal block (8192×3) and rotation block (8192×3). Score block and category block help to determine which points are able to generate grasps. Normal block and rotation block determine the approaching vectors ***n*** and opening vectors ***r*** for all the generated grasps. The approaching distance $d$ for each grasp is computed by following equation:

$$d = \begin{cases} d_{deepest} & \text{if } d_{deepest} < d_{max} \\ d_{max} & \text{if } d_{deepest} \geq d_{max} \end{cases} \quad (4)$$

$d_{deepest}$ is computed by the minimum distance between grasp point $p_i$ and the point clouds under the fingers along approaching vector ***n***. $d_{max}$ is a threshold, which is set to 40mm in our experiment. The losses for score, category, normal and rotation are defined as follows:

$$\mathbf{Loss}_{score} = \frac{1}{M}\sum_i (P^i_{score} - L^i_{score})^2 \cdot m^i_{score} \quad (5)$$

$$\mathbf{Loss}_{category} = \frac{1}{M}\sum_i (L^i_{category} \cdot \ln(P^i_{category}) + (1-L^i_{category}) \cdot \ln(1-P^i_{category}))^2 \cdot m^i_{category} \quad (6)$$

$$\mathbf{Loss}_{normal} = \frac{1}{M}\sum_i (1 - \mathbf{L}^i_{normal} \cdot \mathbf{P}^i_{normal}) \cdot m^i_{normal} \quad (7)$$

$$\mathbf{Loss}_{rotation} = \frac{1}{M}\sum_i (1 - (\mathbf{L}^i_{rotation} \cdot (\mathbf{P}^i_{projection}/\|\mathbf{P}^i_{projection}\|))^2) \cdot m^i_{rotation} \quad (8)$$

, where $\mathbf{P}^i_{projection} = \mathbf{P}^i_{rotation} - (\mathbf{P}^i_{rotation} \cdot \mathbf{L}^i_{normal})\mathbf{L}^i_{normal}$.

$M$ is the number of the input points. $P_{score}$, $P_{category}$, $\mathbf{P}_{normal}$ and $\mathbf{P}_{rotation}$ denote the outputs of score, category, normal and rotation blocks, where $\mathbf{P}_{normal}$ and $\mathbf{P}_{rotation}$ are normalized. For $\mathbf{Loss}_{rotation}$, because $\mathbf{L}^i_{rotation}$ is perpendicular to $\mathbf{L}^i_{normal}$, the projection vector of $\mathbf{P}^i_{rotation}$ on the plane of $\mathbf{L}^i_{rotation}$ is considered: $\mathbf{P}^i_{projection}$. Unlike $\mathbf{Loss}_{normal}$, cosine distance is not suitable to describe the relationship between $\mathbf{P}^i_{projection}$ and $\mathbf{L}^i_{rotation}$, because the rotation prediction $\mathbf{P}^i_{projection}$ has symmetric vector. As shown in Figure 5, the dotted arrow represents the same pose of the gripper with $\mathbf{P}^i_{projection}$, because the gripper has symmetry. However, they will be distinguished by cosine distance. Therefore, we use $1-\cos(x)^2$ to make sure that the losses for symmetric vectors are the same, shown in equation (8). Finally, the total loss is shown in following equation:

$$\mathbf{Loss} = \mathbf{Loss}_{rotation} + \mathbf{Loss}_{normal} + \mathbf{Loss}_{category} + \mathbf{Loss}_{score} \quad (9)$$

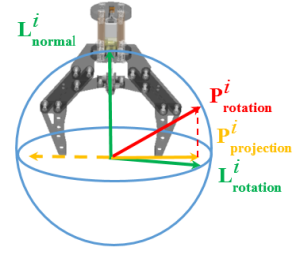

Figure 5. Illustration of relationship between $\mathbf{P}^i_{projection}$ and $\mathbf{L}^i_{rotation}$. The dotted arrow represents symmetric vector of $\mathbf{P}^i_{projection}$.

We use YCB Object Set [26] as our training dataset. Our gripper is shown in Figure 1. For each object model, single-object grasp planning is applied. Then the output are sent to a multi-object grasp planning (Algorithm 1) to construct a cluttered multi-object grasping dataset, which consists of 20k training point clouds, labels and masks. We train our network on 80% of our cluttered multi-object grasping dataset, and keep 20% as a test dataset. Scene models, depth images and supplementary grasps are also recorded for simulation experiments. Our network is trained by stochastic gradient descent using fixed learning rate of $10^{-4}$, momentum of 0.9, and weight decay of $2^{-5}$.

## VI. EVALUATION

### A. Simulation experiments

To do the comparisons with current state-of-art works, we also use YCB objects to train GQCNNs [6] and PointNetGPD [8]. The training parameters and network structures are the same with their papers. We use the test dataset (4k) to do the evaluation. It is worth noting that, grasp sampling methods of GQCNNs and PointNetGPD are based on antipodal grasp sampling and GPG [27] respectively. Moreover, for GQCNNs, it uses cross entropy method (CEM) iteratively to find out grasps with high precision. To generate more grasps for GQCNNs, iteration number is set to 3 and the number of gaussian mixture model is set to one fifth of number of sampling points in our experiments. However, different from these works, our work is an end-to-end generative algorithm with no sampling or search method.

To evaluate the quality of grasps ***g*** generated by different algorithms, their metrics $Q_c(\boldsymbol{g})$ will be calculated using equation (3) combined with objects' models. The grasps with $Q_c(\boldsymbol{g}) > 0$ are regarded as positive labels and vice versa. Moreover, label grasps and their supplementary grasps of test dataset will be searched in generated grasps in order to evaluate the ability of discovering good grasps. Here we use recall-at-high-precision (RAHP) [7], [28] metric to do the evaluations, so thresholds for different algorithms are adjusted until the precision of generated grasps reaches 99%. For our algorithm, we adjust the threshold of category prediction $P_{category}$, which is set to 0.573 in our experiment. The RAHP for label grasps of test dataset is recorded in Figure 6 along with sampling points. It should be emphasized that this recall is not for all the potential grasps but only for label grasps generated by Algorithm 1.

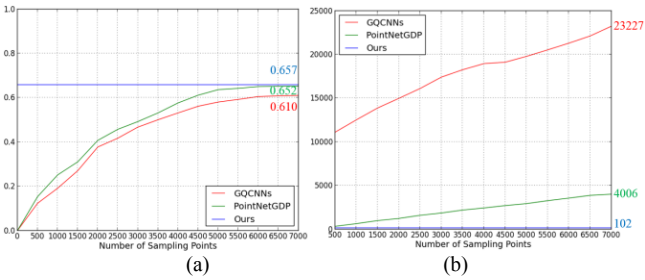

Figure 6. (a) Average recall-at-high-precision(RAHP) curves. (b) Average time cost curves, where unit of y-axis is millisecond. Both of these three algorithms are writen by python and tested in the same computer.

In Figure 6 (a), RAHP is recorded every 500 sampling points. Moreover, the time cost is also recorded in Figure 6. In the figures above, our results are straight lines, because our algorithm has no sampling or search process, which has nothing to do with the number of sampling points or point clouds. Because our test dataset is captured by the camera with random poses, the point clouds can be very sparse. Therefore, PointNetGPD performs better than GQCNNs, because of its 3D feature extractor. Because our algorithm is trained by domain randomization and the network we use is a global feature extractor, the prediction performs better than other algorithms. Some prediction results are shown in Figure 7. From the figures, we find our predicted grasps with high scores are often located near the centroids of objects. Our network can not only regress the poses and scores of all the grasps, but also learn size information of objects. Moreover, as the number of sampling points increases, time cost for these two algorithms will also increase. In our experiment, we find for these two algorithms, especially for GQCNNs, most of time is spent on searching and sampling. Our algorithm is an end-to-end approach and takes only 102ms to finish all the process.

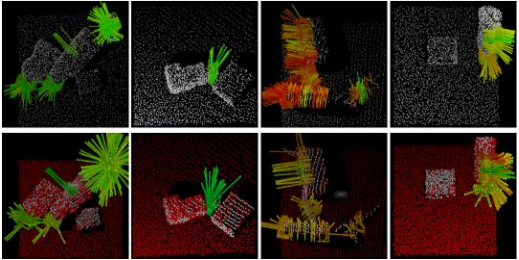

Figure 7. Prediction results of our work for simulation data. The figures in first row are our predictions including grasps and scores. The figures in second row are labels of these point clouds, where the illustration for different colors of points can be found in Figure 4. The colors of grasps from red to green represent predicted scores from low to high.

### B. Robot experiments

We test our algorithm in a UR10 with a low-cost parallel jaw gripper, the open width of which equals to 60 mm. The camera we used is Realsense SR300 with resolution of 640 × 480. To make the real test scene as similar as the simulation environment, several pre-processings are applied for sensor data: 1) The ground plane of point clouds obtained by Realsense is matched and found out using RANSAC; 2) The base coordinate of whole point clouds is transformed on the ground plane, which is to make sure z-axis is perpendicular to the ground plane; 3) The input range is set to: -0.2m<$x$<0.2m, -0.2m<$y$< 0.2m, and point clouds outside this range will be removed. 4) Pre-processings except adding noise in training stage will also be applied for the input. Because of the size limitation of our gripper, not all the objects in YCB set can be grasped. We select 15 objects from them combined with another 15 novel objects, shown in Figure 8. Cartesian planning within MoveIt! framework is applied to control UR10.

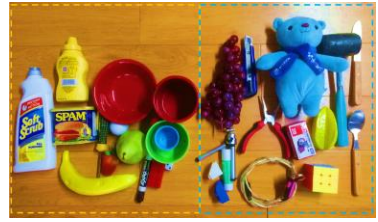

Figure 8. Test objects in real scene. The objects in yellow dashed box are known objects. The objects in bule dashed box are novel objects.

We test 20 times for each algorithm and for each turn, 6 objects are randomly chosen to initialize the test scene. The max number of action attempts for each turn is limited to 10. For GQCNNs and PointNetGPD, their numbers of sampling points are set to the maximum. For our algorithm, grasp with maximum $Q_c$ is executed by the robot. The result is show in following table:

TABLE I. RESULTS OF CLUTTER REMOVAL EXPERIMENTS

| Method | Known objects | | Novel objects | |
|---|---|---|---|---|
| | *SR* | *CR* | *SR* | *CR* |
| PointNetGPD[8] | 78.09% | 94.17% | 67.01% | 84.17% |
| GQCNNs[6] | 74.23% | 92.50% | 65.00% | 83.98% |
| Ours | **82.95%** | **97.50%** | **71.43%** | **91.60%** |

SR and CR denote success rate and completion rate. Success rate is the percentage of successful grasps, while completion rate is the percentage of objects that are removed from the clutter.

In our experiments, our algorithm can get better performance for heavy objects, such as nose plier and hammer, because our network is a global prediction, which helps to locate the centroids of objects. Moreover, the grasp points for some objects, such as bowl and cups are sparse and always suffered by camera noise. For these objects, success rate of our work is higher because our training points are also sparse and the predicted results are generated from all the input points.

## VII. FUTURE WORK

In this paper, an end-to-end approach is proposed to directly predict the poses, categories and scores (qualities) of all the grasps. It takes the whole sparse point clouds as the input and requires no sampling or search process. A fast multi-object grasp detection algorithm based on Ferrari Canny metrics is proposed to generate training dataset. Our experiment shows our work obtains 71.43% success rate and 91.60% completion rate, which performs better than current state-of-art works. In the future, we will consider a better network structure to improve the performance.

## ACKNOWLEDGMENT

Our research has been supported in part by National Natural Science Foundation of China under Grant 61673261.